\documentclass{ifacconf}

\usepackage{graphicx}      
\usepackage{natbib}        
\usepackage{amsmath}
\usepackage{float}
\usepackage{algpseudocode}
\usepackage{algorithm}
\usepackage{booktabs}
\usepackage{subfigure}
\usepackage{subcaption}
\usepackage{subfig}
\usepackage{dsfont}
\usepackage{siunitx}

\begin{document}
\begin{frontmatter}

\title{MP-MPPI: A Motion Primitive Guided Sampling-Based Optimizer for Model Predictive Control} 

\thanks[footnoteinfo]{Corresponding author: {\tt\small marlon.g.mathisen@ntnu.no}}

\author[First]{Marlon G. Mathisen},
\author[First]{Aksel Vaaler},
\author[First]{Olav Egeland},
\author[First]{Eleni Kelasidi}

\address[First]{Norwegian University of Science and Technology, Dept. of Mechanical and Industrial Engineering, Norway} 

\begin{abstract} 
This paper proposes a novel method that extends the Model Predictive Path Integral (MPPI) method with motion primitives for additional structured sampling, which enhances the convergence towards a globally optimal solution. By evaluating motion primitives and perturbed control sequences in a real-time sampling-based optimization loop, this work addresses the limitations of the path planning capabilities of sampling-based controllers. The algorithm is implemented on a quadcopter simulator and tested on an obstacle field navigation task. It is demonstrated that the proposed approach enhances exploration of the control space while maintaining the fast, reactive behavior required for real-time control.

\end{abstract}


\end{frontmatter}

\section{Introduction}
\vspace{-2mm}
The demand for robust, agile controllers has grown significantly in recent years, as researchers pivot to developing field robotic systems that work outside the laboratory, in unpredictable, real-world environments \citep{aljalbout2025realitygaproboticschallenges}. The research community has increasingly been adopting agile quadcopter platforms as a benchmark to evaluate new control algorithms, given the requirements for fast, reactive controllers. Navigating challenging obstacle fields in this setting is particularly demanding, making it well-suited to evaluate novel control strategies \citep{hanover_autonomous_2024}.
While professional quadcopter pilots train extensively to handle these challenging environments \citep{Pfeiffer_2021}, classic path planning approaches often fail to handle the fast-moving, changing environments, and frequently fail to operate in real-time \citep{vehicles3030027}. Therefore, the development of controllers that bridge the gap between high-level planning and low-level control is essential for matching human performance in real-world environments. Model Predictive Control (MPC) bridges this gap by leveraging a system model to predict future states and optimize control inputs \citep{Garcia1989}. The MPC problem is typically solved using gradient-based optimization to satisfy system constraints while minimizing a defined cost function. In recent years, the growing capabilities of parallel computing have enabled the implementation of MPC on high-performance hardware, such as GPUs, allowing for faster and more scalable real-time control in complex robotic systems \citep{8664156}.



\begin{figure}[t]
    \centering
    \includegraphics[width=0.85\linewidth]{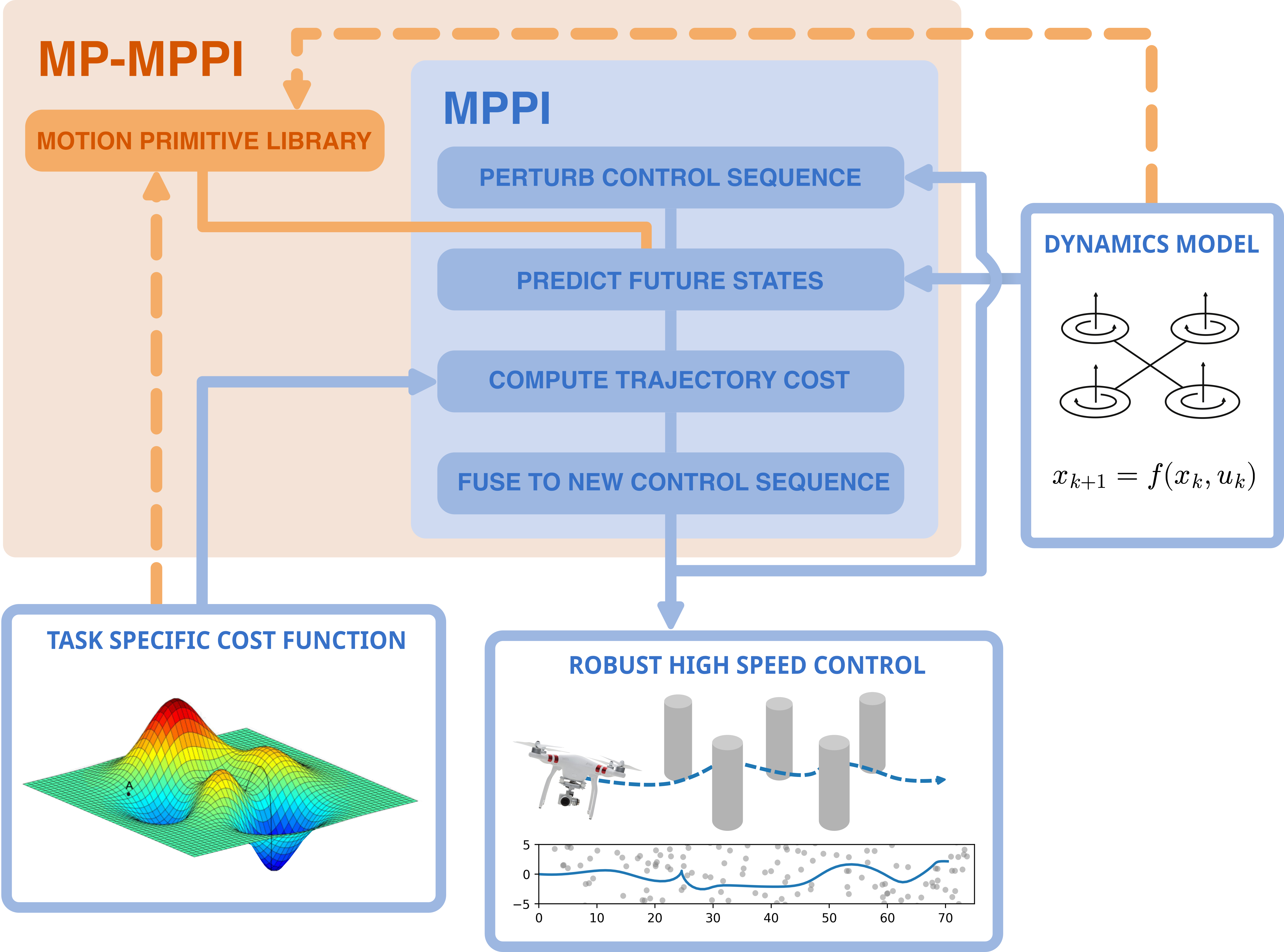}
    \caption{Overview of the proposed MP-MPPI algorithm. The motion
primitives are generated from the cost function and
dynamics model, and are used to inform the model
predictive path integral controller with additional
samples. 
}
    \label{fig:mpmppifigure}
    \vspace{-2mm}
\end{figure}

Model Predictive Path Integral (MPPI) is one such implementation that uses a sampling-based technique for optimization \citep{williams_model_2017}. MPPI works by sampling small perturbations on an initial control input and calculating the cost relating to some control objective for each perturbation. These perturbed control inputs are then fused with a higher weight given to lower cost perturbations, producing a new optimal control sequence. A strength with sampling-based methods is that they do not require differentiable system dynamics, allowing greater freedom to tailor the cost function to the problem at hand. The magnitude of the perturbations added to
the initial control sequence naturally defines a maximum
change in the control sequence per iteration. This limits
how quickly the control input can be transformed into
a new optimal control sequence when the environment
changes, which can make the controller unresponsive and
slow to adapt to new changes. While proven to be a
powerful optimizer, this approach introduces challenges
in balancing exploration and exploitation. Increasing the
magnitude of the perturbations to increase global optimality
often induces instability in the closed-loop system,
resulting in a low-performance controller. 



Several methods have been proposed to increase the global optimality of the MPPI algorithm, while keeping the local optimization capability and cost function design freedoms. DIAL-MPC \citep{xue_full-order_2024} and MPOPI \citep{asmar_model_2023} use the MPPI algorithm iteratively, from high noise to lower noise, to arrive at an optimal control input. This adds to the computation time and lowers the ability to parallelize effectively. The BiC-MPPI method introduced in \cite{jung_bic-mppi_2024} uses the MPPI algorithm bidirectionally to optimize the control input but requires invertible dynamics. \cite{leon_output-sampled_2024} proposed the o-MPPI approach, which changes the algorithm to sample outputs instead of inputs. This modification introduces a limitation, as it also requires invertible dynamics. Earlier work has also attempted to guide the samples with a higher-level planner, such as the RRT guided MPPI introduced in \cite{tao_rrt_2023}. However, in such approaches, the path planners often struggle with real-time performance, which makes replanning more difficult in dynamic environments. Some methods further diversified the parallelization of the MPPI algorithm, such as AERO MPPI \citep{chen_aero-mppi_2025} and U-MPPI \citep{mohamed_towards_2024}. These approaches run multiple parallel instances, optimizing multiple objectives at a time, but this lowers the available samples for each parallel instance, which can lead to lower optimality. Feature-based MPPI introduced in \cite{homburger_feature-based_2022} samples around key features such as emergency braking, to allow quicker decisions in safety-critical situations. However, there is a limit to how many samples each feature can receive, which may reduce performance during normal operation.
Recent approaches, such as Biased-MPPI \citep{trevisan_biased-mppi_2024}, have focused on modifying the sampling distribution of the MPPI algorithm to improve performance, which extends the MPPI to allow arbitrary ancillary controllers to be included in the sampling distribution. \cite{poyrazoglu_c-uniform_2024} proposed the C-Uniformity MPPI method, which modifies the sampling distribution to a uniform distribution. Still, extensively altering the sampling distribution may negatively impact the performance and convergence of the MPPI algorithm, as potentially harmful biases can be introduced from increased variability of sampled inputs.

In this paper, we propose Motion-Primitive-guided Model Predictive Path Integral (MP-MPPI). This is a low-cost, high-impact modification to the MPPI method that increases global optimality and exploration. In this proposed method, additional control sequences are generated as feasible lattice state motion primitives \citep{pivtoraiko_kinodynamic_2011}. These motion primitives serve as informative samples, improving the exploration of the control space. MP-MPPI samples these precomputed motion primitives alongside the standard MPPI perturbations and fuses them to achieve optimal control. An overview of the proposed method can be seen in Figure \ref{fig:mpmppifigure}.

It is shown in the following that the proposed method is: (1) Suited for effective parallel implementation on GPUs. (2) Able to achieve robust control of a simulated quadcopter. (3) Of superior performance compared to the base MPPI algorithm in collision avoidance and obstacle field navigation. 

\vspace{-2mm}
\section{Methodology}
\vspace{-2mm}

The proposed MP-MPPI algorithm was realized by first developing a base MPPI solver. This initial solver was subsequently enhanced through the incorporation of motion primitive samples, resulting in the complete MP-MPPI solver capable of leveraging structured trajectory sampling for improved control performance.

\subsection{Optimal Control Problem}
\vspace{-2mm}
In model predictive control, the following optimal control problem (OCP) is solved \citep{Garcia1989}:
\begin{equation}
\label{OCP}
    \begin{aligned}
    \min_{\{u_t\}_{t=0}^{N-1}} \quad & \sum_{t=0}^{N-1} \ell(x_t, u_t) + \ell_f(x_N) \\
    \text{s.t.} \quad 
    & x_{t+1} = f(x_t, u_t), \quad t = 0, \dots, N-1, \\
    & x_t \in \mathcal{X}, \quad u_t \in \mathcal{U}, \quad t = 0, \dots, N-1, \\
    & x_0 = x_{\text{init}}.
    \end{aligned}
\end{equation}
where the goal is to minimize the cost function \(\ell\) over the prediction horizon \(N\), given an initial system state \(x_{\text{init}}\). The states \(x_t\) and inputs \(u_t\) are constrained by the sets \(\mathcal{X}\) , and \(\mathcal{U}\), as well as needing to satisfy the constraints \(x_{t+1} = f(x_t, u_t)\), where \(f(x_t, u_t)\) is the system dynamics. \(\ell_f\) is the final cost term, which only considers the final predicted state \(x_N\) and can be used as a stabilizing cost to enforce a stable final state.

\subsection{Model Predictive Path Integral}
\vspace{-2mm}
Model Predictive Path Integral (MPPI) is an implementation of the MPC method, where a sampling-based method is used to optimize the control inputs \citep{williams_model_2017}. Samples are generated as perturbed trajectories by modifying the nominal control $u_t$ into the perturbed control variable $u_t^{(k)} = u_t + \epsilon_t^{(k)}$ for $k=1,\ldots, K$ where $\epsilon_t^{(k)}$ is a perturbation that is drawn from $\mathcal{N}(0,\Sigma)$. The perturbed control variables are then used to generate $K$ samples in the form of the perturbed trajectories $x_t^{(k)}$ by the numerical integration scheme
\begin{align}
    x_{t+1}^{(k)} &= x_t^{(k)} + f_{\mathrm{RK4}}(x_t^{(k)}, u_t^{(k)}, \Delta t),
\end{align}
where $f_{\mathrm{RK4}}$ is the Runge-Kutta 4th order integration method. The cost $S^{(k)}$ for each trajectory $x_t^{(k)}$ is defined as:
\begin{align}
    S^{(k)} =  \sum_{t=0}^{N-1} \ell(x_t^{(k)}, u_t^{(k)}) + \ell_f(x_N^j).
\end{align}

The optimal control sequence $u_t$ can then be generated by performing a weighted sum of the perturbed controls $u_t^{(k)}$ following the update law
\begin{align}
\label{MPPI base}
    w_k &= \exp\!\left(-\frac{1}{\lambda} (S^{(k)}-\rho)\right),\\
    \eta &= \sum_{k=1}^{K} \exp\!\left(-\frac{1}{\lambda} (S^{(k)}-\rho)\right),\\
\label{MPPI base end}
    u_t &\leftarrow \sum_{k=1}^{K}\frac{w_k }{\eta}\, u_t^{(k)} ,
\end{align}
where $w_k$ is the weight of the $k$-th perturbed control sequence and $\eta$ is the sum of all weights. The variable \(\rho\) is the minimum of the cost \(S^{(k)}\) over the \(K\) samples, and is included for numerical stability and does not influence the optimality of the solution. The cost \(S^{(k)}\) determines which input sequences are emphasized and which are discarded. The temperature variable \(\lambda\) is a tuning variable that changes the way each perturbation is weighted. A larger \(\lambda\) creates a more uniform weighting, while a smaller \(\lambda\) approaches a hard selection of the least costly perturbed trajectory \citep{williams_model_2017}. The magnitude of the covariance matrix $\Sigma$ defines the rate of adaptation for the optimizer. A lower covariance matrix will lead to more locally optimal results, but decrease the rate of convergence, while a high magnitude covariance matrix increases the convergence speed, but can decrease the stability of the system. 
\vspace{-2mm}
\subsection{MP-MPPI: Motion-Primitive-guided MPPI}
\vspace{-2mm}
This section presents the proposed MP-MPPI extension to the MPPI method, where $N_p$ additional samples in the form of motion primitives are added. 
Motion primitives are sequences of control inputs that represent different types of motion, such as left- and right-turns, changes in velocity, etc. \citep{lavalle_planning_2006}. 

The motion primitive control sequences, denoted $u_t^{(p,j)}$ for the $j$-th motion primtive, produce the trajectories
\begin{equation}
    x_{t+1}^{(p,j)} = f_{\mathrm{RK4}}(x_t^{(p,j)}, u_t^{(p,j)}).
\end{equation}
The control inputs $u_t^{(p,j)}$ are selected from the set of feasible control inputs \(\mathcal{U}\), which ensures the dynamic feasibility of the motion primitives.

State lattice motion primitives are used and are generated by solving the OCP for a change in position to a point reference \citep{pivtoraiko_kinodynamic_2011}. The OCP is solved for a set of $N_p$ point-references, which creates $N_p$ motion primitive samples \(u_t^{(p,0)},\ldots u_t^{(p,N_p-1)}\). These motion primitive samples are combined with white noise samples $u_t^{(0)},\ldots, u_t^{(M-1)}$.

In the optimization loop, $M$ samples $u_\epsilon^0,\ldots,u_\epsilon^{M-1}$ are generated as perturbed control inputs \(u_t^{(k)} = u_t + \epsilon_t^{(k)}\) with white noise $\epsilon^k$. $N_p$ samples are drawn from a library of precomputed motion primitives, denoted $u_t^{(p,j)}$. Each sample is a column of the matrix 
\begin{equation}
    U_t = 
    \begin{bmatrix}
        u_t^{(0)}& 
        \ldots& 
        u_t^{(M-1)}& 
        u_t^{(p,0)} &
        \ldots & 
        u_t^{(p,N_p-1)}
    \end{bmatrix}.
\end{equation}
The cost function is evaluated in parallel for all elements in the vector, to obtain the costs \(S^{(k)}\). Following the weighted sum update scheme from equations \eqref{MPPI base}-\eqref{MPPI base end}, the optimal control sequence is computed as follows:
\begin{equation}
\label{MP step}
   u_t \leftarrow \frac{\sum_{k=1}^{K} \exp\!\left(-\frac{1}{\lambda} (S^{(k)} - \rho)\right) \, U_t^{(k)}}{\sum_{k=1}^{K} \exp\!\left(-\frac{1}{\lambda} (S^{(k)} - \rho)\right)},
\end{equation}
where $K=M+N_p$ and $U^{(k)}$ is column $k$ of $U_t$. 

The full MP-MPPI algorithm can be seen in Algorithm \ref{alg:mp_mppi}.

\begin{algorithm}
\caption{MP-MPPI}
\label{alg:mp_mppi}
\begin{algorithmic}[1]
    \Require
        Current state $x_{\mathrm{curr}}$, 
        previous control sequence $U_{\mathrm{prev}}$ (shape $M\times U_{\mathrm{dim}}$), 
        motion primitives $U_p$ (shape $N_p\times U_{\mathrm{dim}}$),
        horizon $N$, 
        number of noisy samples $M$, 
        noise covariance $\Sigma$, 
        temperature $\lambda$,
        number of motion primitives $N_p$,
        actuator constraints $u_{min}$ and $ u_{max}$
    \Ensure
        Updated control sequence $U_{\mathrm{new}}$

    \State $U_{\mathrm{init}} \leftarrow U_{\mathrm{prev}}$
    \State $x_0 \leftarrow x_{\mathrm{curr}}$

    \Comment{Generate noisy samples}
    \For{$i = 1$ to $M$} 
        \State $\epsilon[i] \sim \mathcal{N}(0,\Sigma)$
        \State $U_{\epsilon}[i] \leftarrow U_{\mathrm{init}} + \epsilon[i]$
    \EndFor

    \State $U_{\epsilon} \leftarrow \text{clip}(U_{\epsilon}, u_{min}, u_{max})$

    \Comment{Construct combined sample set}
    \State $U \leftarrow [\,U_{\epsilon}[0],\dots,U_{\epsilon}[M-1],\,U_p[0],\dots,U_p[N_p-1]\,]$
    \State $K \leftarrow M + N_p$

    \Comment{Evaluate costs}
    \For{$k = 1$ to $K$} 
        \State $\mathrm{cost}[k] \leftarrow \text{RolloutCost}(x_0,\,U[k])$
    \EndFor 
    \State $\rho \leftarrow \displaystyle \min_{k} \mathrm{cost}[k]$
    
    \Comment{Compute weights}
    \For{$k = 1$ to $K$} 
        \State $w[k] \leftarrow \exp\!\Bigl(-\frac{1}{\lambda}\bigl(\mathrm{cost}[k]-\rho\bigr)\Bigr)$
    \EndFor
    
    \State $\eta \leftarrow \sum_{k=1}^{K} w[k]$
    
    \Comment{Update control sequence}
    \State $U_{\mathrm{new}} \leftarrow 0_{N \times U_{\dim}}$
    \For{$k = 1$ to $K$} 
        \State $U_{\mathrm{new}} \leftarrow U_{\mathrm{new}}+ \dfrac{w[k]}{\eta}\,U[k]$
    \EndFor
    \State \Return $U_{\mathrm{new}}$
\end{algorithmic}
\end{algorithm}

\section{Case Study}
\vspace{-3mm}
The proposed method was implemented on a custom quadcopter simulator. The MP-MPPI algorithm was compiled using the JAX Python library and executed on an RTX 2000 Ada Laptop GPU. JAX is a Google-developed, JIT (Just-In-Time) compatible library for array-oriented numerical computations \citep{jax2018github}. This library was used to effectively implement a GPU-vectorized version of the proposed algorithm.
The controller was tuned to run above \SI{100}{\hertz} with satisfactory performance for point reference tracking. To preserve and emphasize the real-time viability of the proposed method, we tuned the controller to be lightweight, based on parameters used in previous real-world MPPI implementations \citep{minarik_model_2024}. The system was simulated using a discretized model with a time step of \SI{10}{\ms}. 

The chosen drone model parameters and the hyperparameters for MPPI are shown in Table \ref{drone model parameters} and Table \ref{mppi param table}, respectively. The controller parameters in Table \ref{mppi param table} are extended to include Table \ref{mp mppi param table} to complete the MP-MPPI controller implementation. The parameters in Table \ref{mp mppi param table} are used to generate the motion primitive library. For each value in the reference lattice, the OCP in (\ref{OCP}) is solved, creating the set of motion primitives, shown in Figure \ref{fig:motion primitives 27}.

\vspace{-2mm}
\subsection{Quadcopter Dynamics Model}
\vspace{-2mm}
Quadcopters are actuated by four propellers, producing forces and torques on the rigid quadcopter body ($u=[f_B, \tau_x, \tau_y, \tau_z]^T$). In this paper, the dynamics model operates with these forces and torques directly. The dynamics model given by \(\Dot{x} = f(x, u)\) was developed based on rigid body equations of motion, with linear dynamics
\begin{align}
\label{linear dynamics1}
    \dot p &= v, \\
\label{linear dynamics2}
    \dot v &= g + \frac{1}{m} R(q)f_B,
\end{align}
which describes linear motion of a system with mass $m$, where $p=[x, y, z]^T$ is the position in the inertial frame, $v$ is the velocity in the inertial frame, $q = [\eta,\sigma^T]^T$ is the unit quaternion of the body orientation and $g$ is the acceleration due to gravity. $f_B$ is the total thrust force in the body frame, which is pre-multiplied by the rotation matrix $R(q)$ to obtain the forces in the inertial frame. Body torques $\tau_B$ produced are given by \([\tau_x, \tau_y, \tau_z\)]. The angular dynamics are described by the equations
\begin{align}
\label{rotation dynamics1}
\dot{q} &= \tfrac{1}{2} q \otimes 
    \begin{bmatrix}
    0 \\
    \omega
    \end{bmatrix} ,\\
\label{rotation dynamics2}
    \dot{\omega} &= J^{-1}(\tau_B - \omega \times (J\omega)),
\end{align}
where $\omega$ is the angular velocity in body coordinates, and \(\tau_B\) is a vector of all body torques. \(J= \mathrm{diag}(I_x, I_y, I_z)\) is the inertia matrix in body coordinates. The inertia matrix is diagonal due to the assumption of symmetry around the center of mass. The quaternion product is given by 
\begin{equation}
    \begin{bmatrix}
        \eta_1\\ \sigma_1
        \end{bmatrix}
        \otimes
        \begin{bmatrix}\eta_2\\\sigma_2
    \end{bmatrix}
    = 
    \begin{bmatrix}
        \eta_1\eta_2 - \sigma_1^T\sigma_2 \\ \
        \eta_1\sigma_2 + \eta_2\sigma_1 + \sigma_1\times\sigma_2
    \end{bmatrix}.
\end{equation}
To enforce these model constraints when calculating the trajectory cost, a 4th order Runge-Kutta method (RK4) was used to integrate the differential equations shown in \eqref{linear dynamics1}-\eqref{rotation dynamics2}. After each integration, the quaternions were normalized to ensure they represented the rotation in unit quaternions.
\subsection{Task Specific Cost Function}
\vspace{-2mm}
The cost function to be minimized was chosen to have a general quadratic form: 
\begin{align}
\label{costfunction}
    \ell(x, v, \omega, q, u) =& c_p \tilde{x}^T \tilde{x} + c_v v^T v  \nonumber\\  &+ c_{\omega} \omega^T \omega + c_{q} d_q(q, q_{\mathrm{ref}})^2 + \tilde{u}^TW\tilde{u}  \nonumber\\ 
    &+ \delta u^T W_\Delta \delta u, \\
    \tilde{x} =& x - x_{\mathrm{ref}}, \\
    \tilde{u} =& u - u_{\mathrm{ref}}, \\
    \delta u  =& u - u_{\mathrm{prev}},
\end{align}
where \(\ell\) is the cost at a specific state. \(x\) is the position, \(v\) is the velocity, \(q\) is the orientation and \(\omega\) is the angular velocity. \(x_{\mathrm{ref}}\) and \(q_{\mathrm{ref}}\) are the commanded references. \(c_p\), \(c_v\), \(c_{q}\) and \(c_{\omega}\) are cost coefficients, \(W\) is a diagonal matrix with input cost coefficients. \(u\) is the input, and \(u_{\mathrm{ref}}\) is the reference input required to sustain a stable hover. $u_{\mathrm{prev}}$ is the previously optimal control sequence, and $W_\Delta$ is a diagonal matrix with change-in-input cost coefficients. The cost coefficients work as regularization terms that penalize high angular and high linear velocity flight. Equally, a rapid change in inputs, or inputs far away from the reference input, can destabilize the controller, and are therefore penalized. The error function for the unit quaternion is given by \citep{minarik_model_2024}
\begin{equation}
    d_q(q, q_{ref}) = 1 - (q \cdot q_{\mathrm{ref}})^{2}, \vspace{2mm} 
\end{equation}
where \([\eta_1, \sigma_1^T]^T\cdot [\eta_2, \sigma_2^T]^T = \eta_1\eta_2 + \sigma_1^T\sigma_2 \) is the quaternion inner product. It is noted \citep{Huynh2009MetricsF3} that ${q \cdot q_{\mathrm{ref}}} = \cos(\tilde{\theta}/2)$ where $\tilde{\theta}$ is the rotation angle of the quaternion deviation $\tilde{q} = \bar{q}_{\mathrm{ref}}\otimes q = [\cos(\tilde{\theta}/2),\sin(\tilde{\theta}/2)\tilde{k}^T]^T$ where $\tilde{k}$ is the unit axis of rotation and $\bar{q}$ is the conjugate of $q$. It follows that $d_q(q, q_{\mathrm{ref}}) = \sin^2(\tilde{\theta}/2)$. The unit quaternion cost term works to encourage upright and forward-facing flight. 

\begin{table}
\centering
\caption{Drone model parameters.}
\label{drone model parameters}
\resizebox{\columnwidth}{!}{
\begin{tabular}{c c c | c c c}
\toprule
\multicolumn{3}{c}{\textbf{UAV model}} & \multicolumn{3}{c}{\textbf{Control limits}} \\
\midrule
Parameter & Units & Value & Parameter & Units & Sim \\
\midrule
$m$     & [kg]        & 1.2   & $F$       & [N]      & $[0.4,\ 36.7]$ \\
$I_x$   & [kg·m$^2$]  & 0.0123  & $\tau_x$  & [N·m]    & $[-1.46,\ 1.46]$ \\
$I_y$   & [kg·m$^2$]  & 0.0123 & $\tau_y$  & [N·m]    & $[-1.46,\ 1.46]$ \\
$I_z$   & [kg·m$^2$]  & 0.0224 & $\tau_z$  & [N·m]    & $[-0.292,\ 0.292]$ \\
\bottomrule
\end{tabular}}
\end{table}

\begin{table}
\centering
\caption{Parameter values used in the MPPI and MP-MPPI controllers.}
\label{mppi param table}
\resizebox{\columnwidth}{!}{
\begin{tabular}{l c | l c | l c}
\toprule
\multicolumn{6}{c}{\textbf{MPPI parameters}} \\
\midrule
Parameter & Value & Parameter & Value & Parameter & Value \\
\midrule
$\Delta t$ [ms]            & 100          & $c_{p}$     & 100.0   & $\Sigma$      & $[1.0,\,0.08,\,0.08,\,0.016]$ \\
$K$                       & 1024          & $c_{q}$     & 50.0    & $W_u$           & $\mathrm{diag}(0.01,\,0.05,\,0.05,\,0.10)$ \\
$N$                       & 15           & $c_{v}$     & 10.0    & $W_\Delta$  & $\mathrm{diag}(0.5,\,1,\,1,\,5)$ \\
$\lambda$                 & $10^{4}$    & $c_{\omega}$& 0.05    &       $c_{\text{obs}}$      &  $10^9$\\
                            &           &              &         &         $d_{\text{safe}}$      &   $0.2$ \\
\bottomrule
\end{tabular}}
\end{table}

\begin{table}
\centering
\caption{Parameter values used to generate the set of motion primitives.}
\label{mp mppi param table}
\resizebox{\columnwidth}{!}{
\begin{tabular}{l c | l c | l c}
\toprule
\multicolumn{6}{c}{\textbf{Motion Primitive Parameters}} \\
\midrule
Parameter & Value & Parameter & Value & Parameter & Value \\
\midrule
$\text{ref}_x$     & [-5, 0, 5]   & $\text{ref}_y$     & [-5, 0, 5]   & $\text{ref}_z$     & [-5, 0, 5] \\
\bottomrule
\end{tabular}}
\end{table}

\begin{figure}[ht!]
    \centering
    \vspace{-5mm}
    \includegraphics[width=0.73\linewidth]{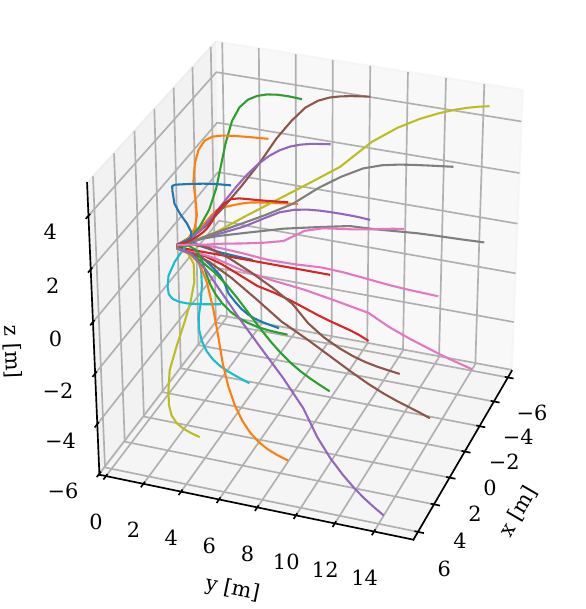}
    \vspace{-2mm}
    \caption{Resulting trajectories from executing the motion primitives that minimize the cost to point references with the parameters in Table \ref{mp mppi param table}. The quadcopter is traveling at \SI{7}{\metre\per\second} in the positive y direction.}
    \label{fig:motion primitives 27}
\end{figure}
\subsection{Obstacle Avoidance}
\vspace{-2mm}
Obstacle avoidance is achieved through a cost modification of \eqref{costfunction}. Adding high cost to the trajectories that lead to collision means that those trajectories are effectively disregarded when calculating the updated control sequence in \eqref{MPPI base}. The new trajectory cost becomes
\begin{equation}
S^{(k)}{'} = S^{(k)} + \sum_{j=1}^{N_{\mathrm{obs}}} c_{\text{obs}} \cdot \mathds{1}(x_j^k \in C_{\text{obs}}),
\label{eq:avoidance_cost}
\end{equation}

where \(c_{obs}\) is a penalty for colliding, $N_{\mathrm{obs}}$ is the number of obstacles and \(\mathds{1}\) is the indicator function that returns 1 when \(x_j^k \in C_{\text{obs}}\) and 0 otherwise. This function assumes that the obstacle geometry is accessible. To account for the physical dimensions of the drone, a safety distance $d_{\text{safe}}$ is included in the collision check.
\section{Results}
\vspace{-2mm}
\subsection{Obstacle Field Navigation}
\begin{figure}[ht!]
    \centering
    \includegraphics[width=\linewidth]{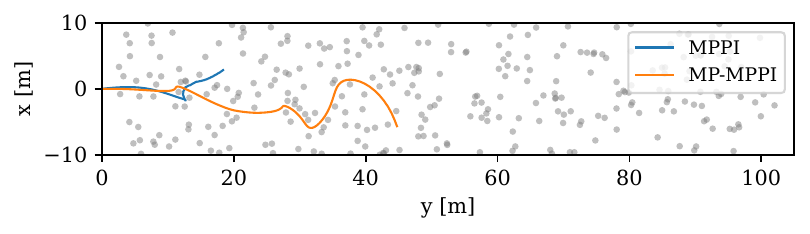}
    \vspace{-5mm}
    \caption{Example of the obstacle field navigation task illustrating how the MP-MPPI algorithm explores further and navigates the obstacle field more effectively.}
    \label{fig:navigation_task}
\end{figure}

\vspace{-2mm}
Obstacle avoidance and navigation are evaluated on a dense obstacle field navigation task. An example is shown in Figure \ref{fig:navigation_task}. The quadcopter has to navigate through 300 randomly placed pillars with a diameter of \SI{1}{\metre}, generated within a bounding box of \SI{20}{\metre} by \SI{100}{\metre}. The drone is commanded to follow a moving point reference \SI{10}{\metre} in front of the quadcopter on the center line of the obstacle field. Each simulation lasted for \SI{5}{\second} and started from a hovering state. The obstacle field navigation task was run 100 times, with the end positions, as well as the number of collisions, being recorded. 

\begin{figure}
    \centering
    \includegraphics[width=0.75\linewidth]{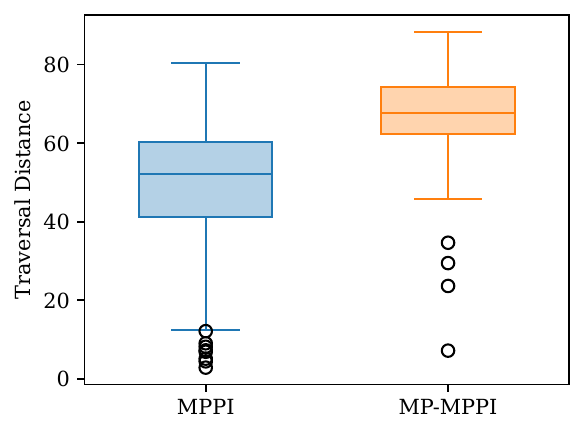}%
    \vspace{-2mm}
    \caption{Box plot showing the difference in performance between the MPPI and MP-MPPI algorithms.}
    \label{fig:traversal_comparison}
\end{figure}

As seen in Figure \ref{fig:traversal_comparison}, the MPPI algorithm navigated the obstacle field effectively with a mean traversal distance of \SI{48.0}{\metre} and with a standard deviation of \SI{18.0}{\metre}. Adding 27 motion primitives (visualized in Figure \ref{fig:motion primitives 27}) to create the MP-MPPI controller, increased the mean traversal distance to \SI{66.6}{\metre}, with a standard deviation of \SI{12.6}{\metre}. The MPPI controller collided once during the 100 runs, while the MP-MPPI controller managed to avoid any collisions. This demonstrated that the addition of a few motion primitives improved the obstacle field navigation in both total distance traversed and in collisions avoided. The MP-MPPI algorithm discovers trajectories that take it beyond its locally minimum solution, and further progress through the obstacle field.

\subsection{Single Update Analysis}
\begin{figure}
    \centering
    \includegraphics[width=1\linewidth]{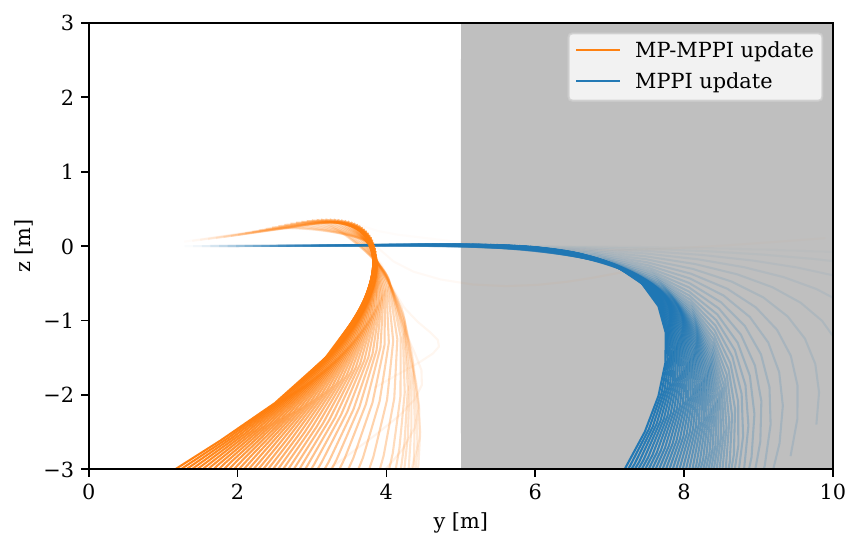}
    \vspace{-5mm}
        \caption{Trajectory updates for 50 MPPI and MP-MPPI iterations when approaching a wall at high speed (\SI{12}{\metre\per\second}) from a close distance (\SI{5}{\metre}). The faded trajectories show previous optimal trajectories, while hard-drawn lines show more recent optimal trajectories.}

    \label{fig:wall}
\end{figure}

\vspace{-2mm}
The MP-MPPI algorithm and the MPPI base algorithm are evaluated on reactivity, where both algorithms are compared in a scenario where the drone is moving towards a wall at \SI{10}{\metre\per\second} from a distance of \SI{5}{\metre}, and suddenly becomes aware of the obstacle. The controller performance can be seen in Figure \ref{fig:wall}.

The MP-MPPI algorithm generates a trajectory that avoids collision with the wall, while the MPPI algorithm collides with it. This highlights how the motion primitive samples go into effect when the base MPPI algorithm is limited by the magnitude of the perturbations. The additional motion primitives inform the controller of solutions that might not be locally optimal, but represent the needed motion to avoid collision with the wall.

\subsection{Controller Update Frequency}
\begin{figure}
    \centering
    \includegraphics[width=0.9\linewidth]{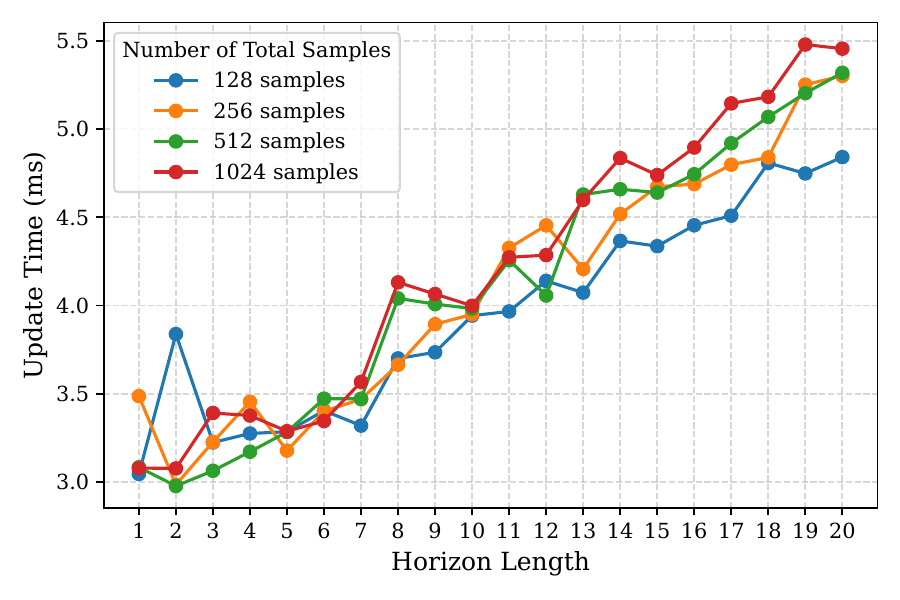}
    \caption{Latency of online optimization step used in MP-MPPI/MPPI for $M$ samples, and $N$ step long prediction horizon.}
    \label{fig:update_frequencies} 
\end{figure}

Agile robotic systems rely on rapid feedback for stable control, therefore the update frequency is also benchmarked. Figure \ref{fig:update_frequencies} shows the update frequencies for a varying number of samples and prediction horizons. This shows that the computation times follow a linear trend with respect to the horizon length. It also shows that the algorithm effectively runs on a GPU, with only small differences in latency between using 1024 samples and 128 samples. This allows for the evaluation of additional motion primitive samples, with a limited effect on the update frequency.

\subsection{Discussion and Limitations}


While the proposed MP-MPPI method demonstrates that a small selection of precomputed motion primitives can significantly increase the performance over the base MPPI algorithm, the extent of this effect has not been fully explored. While the convergence has been proven for arbitrary controllers \citep{trevisan_biased-mppi_2024}, there is still the need to explore if these kinematically feasible building blocks provide better convergence and introduce less bias than an arbitrary controller would.

A main limitation of this approach is that the balance between motion primitives and regular noisy samples becomes a design parameter that must be carefully tuned. Too many motion primitives can cause the controller to weight potentially suboptimal trajectories, thereby lowering performance and optimality. Future work may investigate if there is a method to generate the motion primitive library for specific tasks optimally.

\section{Conclusion}

In this work, we propose MP-MPPI, an enhanced version of the MPPI algorithm that improves global optimality by introducing exploratory motion primitives into the MPPI update law. The performance is evaluated in a quadcopter simulator, which indicated the method's ability to navigate highly cluttered environments. Obtained results demonstrated that in an unpredictable environment, MP-MPPI can adapt quickly and avoid collision, where the MPPI fails to do so.

The proposed MP-MPPI method is shown to be a low-cost, high-impact extension to the MPPI method, capable of running on lower-end parallel computing hardware, and thus it may serve as a low-cost method of achieving robustness in autonomous systems. 

\bibliography{ifacconf}

@article{8664156,
 author = {Abughalieh, Karam M. and Alawneh, Shadi G.},
 doi = {10.1109/ACCESS.2019.2904240},
 journal = {IEEE Access},
 number = {},
 pages = {34348-34360},
 title = {A Survey of Parallel Implementations for Model Predictive Control},
 volume = {7},
 year = {2019}
}

@misc{aljalbout2025realitygaproboticschallenges,
author = {Aljalbout, Elie and Xing, Jiaxu and Romero, Angel and Akinola, Iretiayo and Garrett, Caelan and Heiden, Eric and Gupta, Abhishek and Hermans, Tucker and Narang, Yashraj and Fox, Dieter and Scaramuzza, Davide and Ramos, Fabio},
year = {2025},
month = {10},
pages = {},
title = {The Reality Gap in Robotics: Challenges, Solutions, and Best Practices},
doi = {10.48550/arXiv.2510.20808}
}

@misc{asmar_model_2023,
 author = {Asmar, Dylan M. and Senanayake, Ransalu and Manuel, Shawn and Kochenderfer, Mykel J.},
 doi = {10.48550/arXiv.2203.16633},
 month = {March},
 publisher = {arXiv},
 title = {Model {Predictive} {Optimized} {Path} {Integral} {Strategies}},
 year = {2023}
}

@misc{chen_aero-mppi_2025,
 author = {Chen, Xin and Huang, Rui and Tang, Longbin and Zhao, Lin},
 doi = {10.48550/arXiv.2509.17340},
 month = {September},
 publisher = {arXiv},
 title = {{AERO}-{MPPI}: {Anchor}-{Guided} {Ensemble} {Trajectory} {Optimization} for {Agile} {Mapless} {Drone} {Navigation}},
 year = {2025}
}

@article{Garcia1989,
 author = {Garc{\'\i}a, C.\,E. and Prett, D.\,M. and Morari, M.},
 doi = {10.1016/0005-1098(89)90002-2},
 journal = {Automatica},
 number = {3},
 pages = {335--348},
 title = {Model predictive control: Theory and practice --- {A} survey},
 volume = {25},
 year = {1989}
}

@article{hanover_autonomous_2024,
 author = {Hanover, Drew and Loquercio, Antonio and Bauersfeld, Leonard and Romero, Angel and Penicka, Robert and Song, Yunlong and Cioffi, Giovanni and Kaufmann, Elia and Scaramuzza, Davide},
 doi = {10.1109/TRO.2024.3400838},
 journal = {IEEE Transactions on Robotics},
 pages = {3044--3067},
 title = {Autonomous {Drone} {Racing}: {A} {Survey}},
 volume = {40},
 year = {2024}
}

@article{homburger_feature-based_2022,
 author = {Homburger, Hannes and Wirtensohn, Stefan and Diehl, Moritz and Reuter, Johannes},
 doi = {10.3390/machines10100900},
 journal = {Machines},
 month = {October},
 number = {10},
 pages = {900},
 title = {Feature-{Based} {MPPI} {Control} with {Applications} to {Maritime} {Systems}},
 volume = {10},
 year = {2022}
}

@article{Huynh2009MetricsF3,
author = {Huynh, Du},
year = {2009},
month = {10},
pages = {155-164},
title = {Metrics for 3D Rotations: Comparison and Analysis},
volume = {35},
journal = {Journal of Mathematical Imaging and Vision},
doi = {10.1007/s10851-009-0161-2}
}

@misc{jung_bic-mppi_2024,
 author = {Jung, Minchan and Kim, Kwangki},
 doi = {10.48550/arXiv.2410.06493},
 month = {October},
 publisher = {arXiv},
 title = {{BiC}-{MPPI}: {Goal}-{Pursuing}, {Sampling}-{Based} {Bidirectional} {Rollout} {Clustering} {Path} {Integral} for {Trajectory} {Optimization}},
 year = {2024}
}

@book{lavalle_planning_2006,
 author = {LaValle, Steven M.},
 doi = {10.1017/CBO9780511546877},
 month = {May},
 publisher = {Cambridge University Press},
 title = {Planning {Algorithms}},
 year = {2006}
}

@inproceedings{leon_output-sampled_2024,
  title={Output-sampled model predictive path integral control (o-MPPI) for increased efficiency},
  author={Yan, Leon Liangwu and Devasia, Santosh},
  booktitle={2024 IEEE International Conference on Robotics and Automation (ICRA)},
  pages={14279--14285},
  year={2024},
  organization={IEEE}
}

@misc{minarik_model_2024,
 author = {Minarik, Michal and Penicka, Robert and Vonasek, Vojtech and Saska, Martin},
 doi = {10.48550/arXiv.2407.09812},
 month = {July},
 publisher = {arXiv},
 title = {Model {Predictive} {Path} {Integral} {Control} for {Agile} {Unmanned} {Aerial} {Vehicles}},
 year = {2024}
}

@misc{mohamed_towards_2024,
 author = {Mohamed, Ihab S. and Xu, Junhong and Sukhatme, Gaurav S. and Liu, Lantao},
 doi = {10.48550/arXiv.2306.12369},
 month = {December},
 publisher = {arXiv},
 title = {Towards {Efficient} {MPPI} {Trajectory} {Generation} with {Unscented} {Guidance}: {U}-{MPPI} {Control} {Strategy}},
 year = {2024}
}

@article{Pfeiffer_2021,
 author = {Pfeiffer, Christian and Scaramuzza, Davide},
 doi = {10.1109/lra.2021.3064282},
 journal = {IEEE Robotics and Automation Letters},
 month = {apr},
 number = {2},
 pages = {3467-3474},
 publisher = {Institute of Electrical and Electronics Engineers (IEEE)},
 title = {Human-Piloted Drone Racing: Visual Processing and Control},
 volume = {6},
 year = {2021}
}

@inproceedings{pivtoraiko_kinodynamic_2011,
 author = {Pivtoraiko, Mihail and Kelly, Alonzo},
 booktitle = {2011 {IEEE}/{RSJ} {International} {Conference} on {Intelligent} {Robots} and {Systems}},
 doi = {10.1109/IROS.2011.6094900},
 month = {September},
 pages = {2172--2179},
 title = {Kinodynamic motion planning with state lattice motion primitives},
 year = {2011}
}

@misc{poyrazoglu_c-uniform_2024,
 author = {Poyrazoglu, O. Goktug and Cao, Yukang and Isler, Volkan},
 doi = {10.48550/arXiv.2409.12266},
 month = {September},
 publisher = {arXiv},
 title = {C-{Uniform} {Trajectory} {Sampling} {For} {Fast} {Motion} {Planning}},
 year = {2024}
}

@misc{tao_rrt_2023,
 author = {Tao, Chuyuan and Kim, Hunmin and Hovakimyan, Naira},
 doi = {10.48550/arXiv.2301.13143},
 month = {January},
 publisher = {arXiv},
 title = {{RRT} {Guided} {Model} {Predictive} {Path} {Integral} {Method}},
 year = {2023}
}

@article{trevisan_biased-mppi_2024,
 author = {Trevisan, Elia and Alonso-Mora, Javier},
 doi = {10.1109/LRA.2024.3397083},
 journal = {IEEE Robotics and Automation Letters},
 month = {June},
 number = {6},
 pages = {5871--5878},
 title = {Biased-{MPPI}: {Informing} {Sampling}-{Based} {Model} {Predictive} {Control} by {Fusing} {Ancillary} {Controllers}},
 volume = {9},
 year = {2024}
}

@article{vehicles3030027,
 author = {Karur, Karthik and Sharma, Nitin and Dharmatti, Chinmay and Siegel, Joshua E.},
 doi = {10.3390/vehicles3030027},
 journal = {Vehicles},
 number = {3},
 pages = {448--468},
 title = {A Survey of Path Planning Algorithms for Mobile Robots},
 volume = {3},
 year = {2021}
}

@article{williams_model_2017,
 author = {Williams, Grady and Aldrich, Andrew and Theodorou, Evangelos A.},
 doi = {10.2514/1.G001921},
 journal = {Journal of Guidance, Control, and Dynamics},
 month = {February},
 number = {2},
 pages = {344--357},
 title = {Model {Predictive} {Path} {Integral} {Control}: {From} {Theory} to {Parallel} {Computation}},
 volume = {40},
 year = {2017}
}

@misc{xue_full-order_2024,
 author = {Xue, Haoru and Pan, Chaoyi and Yi, Zeji and Qu, Guannan and Shi, Guanya},
 doi = {10.48550/arXiv.2409.15610},
 month = {September},
 publisher = {arXiv},
 title = {Full-{Order} {Sampling}-{Based} {MPC} for {Torque}-{Level} {Locomotion} {Control} via {Diffusion}-{Style} {Annealing}},
 year = {2024}
}

@software{jax2018github,
  author = {James Bradbury and Roy Frostig and Peter Hawkins and Matthew James Johnson and Chris Leary and Dougal Maclaurin and George Necula and Adam Paszke and Jake Vander{P}las and Skye Wanderman-{M}ilne and Qiao Zhang},
  title = {{JAX}: composable transformations of {P}ython+{N}um{P}y programs},
  url = {http://github.com/jax-ml/jax},
  version = {0.3.13},
  year = {2018},
}

\end{document}